\documentclass{amia}
\usepackage{lipsum} 
\usepackage{bm}
\usepackage{times}
\usepackage{helvet}
\usepackage{courier}
\usepackage{times}
\usepackage{latexsym}
\usepackage{amsmath}
\usepackage{algorithm}
\usepackage{algpseudocode}
\usepackage{url}
\usepackage{multirow}
\usepackage{enumitem}
\usepackage{array}
\usepackage[symbol]{footmisc}
\usepackage{threeparttable}
\usepackage{graphicx}
\usepackage{dblfloatfix}
\usepackage{caption}
\usepackage[labelformat=simple]{subcaption}

\usepackage{adjustbox}
\usepackage{multirow}
\usepackage{hyperref}
\usepackage{xcolor}
\usepackage{float}
\usepackage{longtable}
\usepackage{pdflscape}
\usepackage{rotating}

\usepackage{amsmath}
\usepackage{amssymb}
\usepackage{subcaption}
\usepackage{tabularx}
\usepackage{makecell}
\usepackage{mathtools}

\newcolumntype{P}[1]{>{\centering\arraybackslash}p{#1}}

\usepackage{import}

\usepackage{natbib}
\setcitestyle{authoryear,open={(},close={)}}

\usepackage[toc,page]{appendix}

\begin{document}

\title{Large Language Model-based Role-Playing for Personalized Medical Jargon Extraction}



\author{
Jung Hoon Lim, MSc$^{1}$, Sunjae Kwon,  MSc$^{1}$, Zonghai Yao, MSc$^1$, John P. Lalor, PhD$^2$, Hong Yu, PhD$^{1,3,4,5}$
}

\institutes{
    $^1$ Manning College of Information and Computer Sciences, University of Massachusetts Amherst, MA, USA\\
    $^2$ Mendoza College of Business, University of Notre Dame, IN, USA\\
    $^3$ Department of Medicine, University of Massachusetts Medical School, Worcester, MA, USA\\
    $^4$ Center for Biomedical and Health Research in Data Sciences, Miner School of Computer and Information Sciences, University of Massachusetts Lowell, MA, USA\\
    $^5$ Center for Healthcare Organization and Implementation Research, VA Bedford Health Care, MA, USA\\
}

\maketitle

\textbf{Corresponding author:} Hong Yu, PhD 

Center for Biomedical and Health Research in Data Sciences, Manning College of Computer and Information Sciences, University of Massachusetts Lowell, MA, USA 

Phone: 1 978-934-3620  
Email: Hong\_Yu@uml.edu

\section{Abstract}
\textbf{Objective:}
Studies reveal that Electronic Health Records (EHR), which have been widely adopted in the U.S. to allow patients to access their personal medical information, do not have high readability to patients due to the prevalence of medical jargon.
Tailoring medical notes to individual comprehension by identifying jargon that is difficult for each person will enhance the utility of generative models.
In this study, we investigate whether role-playing in a large language model, ChatGPT, can improve performance on customized medical term extraction based on socio-demographic groups.

\textbf{Materials and Methods:} 
We gathered data on extracting medical terms from 270 Mechanical Turk workers, two open-source biomedical NER systems (SciSpacy and MedJEx), and ChatGPT, using 20 sentences selected from medical notes.
Each medical term was labeled based on the consensus of the 270 Mechanical Turk respondents. 
The performance of ChatGPT in extracting medical terms was compared to that of earlier models.
Ultimately, we assessed the effect of role-playing with and without in-context learning in ChatGPT quantitatively with 14 different socio-demographic groups and varying temperatures between 0.0 and 1.0 and calculated the F1 scores.

\textbf{Results:} 
Without in-context learning, F1 scores improved in 133 out of 140 cases with role-playing.
Wilcoxon signed-rank test further shows that the F1 scores follow distinct distributions.
The application of in-context learning, with four examples each, resulted in significant performance improvements while being less affected by role-playing.
Particularly, GPT4, with in-context learning and role-playing, gave a macro-averaged F1 score of 51.28, showing better performance than the previous state-of-the-art model, MedJEx, which scored 50.92.

\textbf{Conclusion and Future Work:} 
The results show that role-playing in ChatGPT can give personalized results for medical term extraction.
We intend to explore experiments further to harness the potential of role-playing in ChatGPT for practical applications such as EHR note personalization, medical concept linking systems, or chatbot-based self-diagnosis systems.



\section{Introduction}

Medical notes are vital documents in healthcare, providing crucial information about patients' conditions, treatments, and medical history.
However, their utility can be impeded by the presence of domain-specific jargon, which can hinder comprehension, particularly for individuals without specialized medical training.
Studies reveal that Electronic Health Records (EHR), which have been widely adopted in the U.S. to allow patients to access their personal medical information, do not have high readability to patients due to the prevalence of medical jargon \citep{chen2018natural}.
Addressing this issue by tailoring medical notes to individual comprehension levels through the identification of challenging medical terms has the potential to enhance the effectiveness of generative models.

There have been studies to identify medical terms in texts.
Biomedical Named Entity Recognition (BioNER), which is a task for identifying biomedical named entities from medical text, have been actively studied \citep{soomro2017bio, cariello2021comparison, khan2020mt}.
Another related task is Key Phrase Extraction in the medical domain \citep{sarkar2013hybrid, gero2019namedkeys}.
Furthermore, \citep{kwon2022MedJEx} considered the reader's comprehension of each medical jargon.
However, these methods do not consider the reader's personal information, such as their socio-demographic background.

The advance of large language models (LLMs) has brought many emerging behaviors that were not present in smaller models \citep{wei2022emergent}.
These models can be used to serve as agents that simulate human behaviors in various roles, as directed through input prompts.
However, as far as we know, there are no results that quantify LLMs' performance in role-playing in the medical domain.

In this study, we investigate whether role-playing in a large language model, ChatGPT, can improve performance on medical term extraction based on socio-demographic groups.
In particular, we use the result of 270 Amazon Mechanical Turk workers' (hereafter "Turkers") work in extracting medical jargon from 20 sentences.
These workers possess diverse sociodemographic backgrounds, spanning various group types including age, education level, frequency of health literacy reading, and gender.
Considering the majority vote of Turkers as the ground truth label, our experiments show that ChatGPT's prediction on each group improves in most cases when role-playing is applied.
Furthermore, the results reveal that, with in-context learning, GPT-4 outperforms MedJEX, the state-of-the-art model for medical term extraction.

Our contributions are as follows:
\begin{itemize}
    \item We present the first quantitative analysis to measure the impact of role-playing of LLM in the medical domain. We experiment with and without in-context learning in ChatGPT on reflecting the perspectives of individuals from diverse socio-demographic backgrounds.
    \item We evaluated ChatGPT's performance in medical term extraction and compared it with the performance of previous models.
    \item Our research demonstrated ChatGPT's capability to enhance traditional BioNER systems by using role-play to achieve personalized patient education, a potential not previously met by existing models.
\end{itemize}

\begin{figure*}[htbp]
    \centering
    \includegraphics[width=\linewidth]{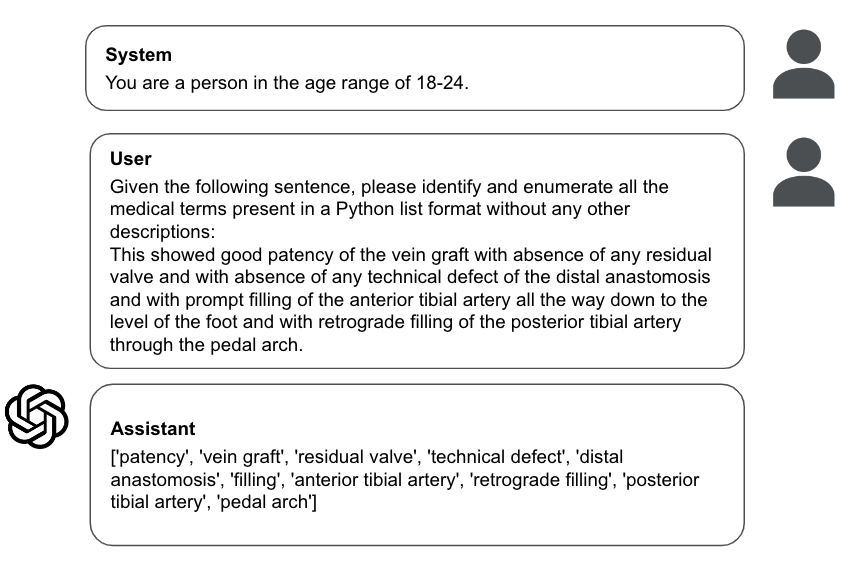}
    \caption{Message interactions with ChatGPT. A system message is sent to set the ChatGPT assistant's role. Then, a user message that contains the query is inputted such that ChatGPT could give the response in the requested format (Python list).}
    \label{fig:MCC_difference}
\end{figure*}

\section{Related Work}

\subsection*{Large Language Models (LLM) and ChatGPT}

The objective of the task of language modeling is to predict a series of words based on the corpora which a model is trained with.
Transformer language models \citep{vaswani2017attention} trained on large corpora of unlabeled texts have shown to be effective on general-purpose semantic features.
Specifically, pre-trained language models (PLMs) based on Transformer could easily adapt to downstream tasks after being fine-tuned on a small amount of labeled data.

Scaling up PLMs in a number of parameters and data size often leads to a better performance on downstream tasks \citep{radford2019language, brown2020language, chowdhery2022palm}.
A continued trend of scaling up PLMs has resulted in the emergence of Large Language Models (LLMs), which refer to language models with billions of parameters trained on massive amounts of data.
These LLMs have shown to have ``emergent behaviors'' \citep{wei2022emergent} which are behaviors not present in smaller models.
Some examples of these ``emergent behaviors'' are in-context learning, instruction-following, and step-by-step reasoning.

In-context learning ability, which is formally introduced by GPT-3 \citep{brown2020language}, enables a language model to perform a task without additional training through a given instruction and/or several examples as a prompt.
Instruction following refers to a behavior where large language models improve on tasks based on instructions after being fine-tuned with a dataset consisting of (INSTRUCTION, OUTPUT) pairs \citep{wei2021finetuned, ouyang2022training, sanh2021multitask}. 
Step-by-step reasoning is a process where LLMs solve complex tasks through a chain-of-thought prompting strategy, which uses a prompting mechanism with a series of intermediate reasoning steps to generate the final output \citep{wei2022chain}.

Generative Pre-trained Transformer (GPT) is a type of large language model that uses decoder-only Transformer models to predict the next word or token.
The GPT series began in 2018 with GPT-1 \citep{radford2018improving} which has 0.18 billion parameters.
The series has progressively scaled up, resulting in increasingly powerful models.
GPT-2 had 1.5 billion parameters and was trained with a goal of performing tasks via unsupervised language modeling without further fine-tuning.
GPT-3 once again gave a big performance gap with 175 billion parameters and showing emergent behaviors such as in-context learning.
The capacity of GPT-3 could be further enhanced through training on code data \citep{fu2022does} and applying reinforcement learning from human feedback \citep{ouyang2022training}, resulting in the creation of GPT-3.5.
Most recently, GPT-4 was introduced with 1.7 trillion parameters with better performance in many evaluation tasks \citep{openai2023gpt4, bubeck2023sparks}.
GPT-4 was trained with a six-month iterative alignment with safety reward and a number of intervention strategies have been applied to mitigate potential harms such as generating hallucinations or inappropriate responses.

ChatGPT is a conversational model made based on GPT models released on November 2022 by OpenAI \citep{Gertner_2023}.
ChatGPT provides Application programming interfaces (APIs) which allow us to have relatively unconstrained access to fine-tune and infer from powerful LLMs, giving us huge possibilities of its usage. 

\subsection*{Role Playing and Personalizing in LLMs}
Role-playing in LLMs refers to when LLMs are prompted to play a specific role.
Prior work \citep{shanahan2023role} foregrounds the concept of role-playing as the models that appear to act like a human being, in fact, simply generating the next token or word based on previous texts.
It deals with cases such as deception and self-awareness and suggests that a dialogue agent can be seen as a superposition of simulacra that varies based on previous texts. 
\citep{li2023chatharuhi} exploits the in-context learning abilities of LLMs to mimic the conversational styles of fictional characters by providing example passages of each character.
Recent research has shown LLMs with various assigned roles to cooperate in a multi-agent setting to perform tasks \citep{chen2024persona, chen2023autoagents, li2023camel, wei2023multi, wang2023unleashing}.
\citep{gerosa2024can} discusses if AI and LLMs can replace human responses and behaviors in research settings in persona-based prompting for interviews, multi-persona dialogue for focus groups, and mega-persona responses for surveys.


Personalization in LLMs focuses on tailoring model outputs to individual users, considering both benefits and risks at the individual and societal levels \citep{kirk2023personalisation}.
An application of personalization in LLMs is item recommendations.
\citep{lyu2023llm} uses diverse prompts and input augmentation for LLMs to improve on generating recommendations for items.
It leverages user engagement data to identify important neighbor items.
\citep{chen2023palr} fine-tunes an LLM to retrieve and recommend items based on user history and a list of candidate items.
\citep{salemi2023lamp} proposes a retrieval augmentation solution where personalized items that will be included into the instructions are retrieved from user profile for each input.
Another key application of personalization in LLMs is content generation. 
\citep{deshpande2024proceedings} explores various approaches in this area. \citep{alhafni2024personalized} presents a novel benchmark for training generative models to personalize text by controlling fine-grained linguistic attributes, systematically evaluating the performance of various large language models and providing insights into factors affecting their performance. 
\citep{oh2024use} examines the use of language models for personalized synthetic text generation in the biomedical domain, particularly focusing on medical records and evaluating the impact of the LM’s knowledge and size alongside objective and clinical assessments. 
Additionally, \citep{Tran2024ReadCtrlPT} introduces the ReadCtrl method, which instruction-tunes LLMs to tailor users' readability levels for personalized content generation.

\subsection*{Medical Term Extraction}

The goal of medical term extraction is to help improve patients' comprehension of their EHR notes by identifying the terms that may be unfamiliar to patients.
The task is different from BioNER, which aims to recognize and classify biomedical entities from text without considering patients' comprehension \citep{soomro2017bio, cariello2021comparison, khan2020mt}.

Another related task is key phrase extraction, which identifies important phrases or clauses that represent topics \citep{sarkar2013hybrid, gero2019namedkeys}. SciSpacy \citep{neumann-etal-2019-SciSpacy} is an NLP library made to satisfy the primary text processing needs in the biomedical domain. It is built on the SpaCy \citep{honnibal2017spacy} library with models retrained for POS tagging, dependency parsing, and named entity recognition. 
MedJEx \citep{kwon2022MedJEx} and README \citep{yao2023readme} were specifically designed to identify medical jargon terms in electronic health record notes and provide their corresponding lay definitions.
It is trained on hyperlink spans of Wikipedia articles and fine-tuned on the annotated MedJ dataset, which consists of 18,178 sentences annotated by domain experts.
It further elicits UMLS concepts using QuickUMLS \citep{soldaini2016quickumls}, which is used to extract binary features with their weighting scores calculated via term-frequency scores and masked language model scores.


\section{Dataset}
The 20 sentences were randomly selected from the PittEHR database \citep{Voorhees2013-cv}, which includes de-identified patient records. If sentences were shorter than ten words or only included administrative data, they were discarded and sampling continued until we reached 20. These sentences consist of 904 words or 709 terms.


Medical jargons often consist of consecutive words that, when separated, would not individually qualify as medical jargon.
Consequently, we divided each sentence into a group of terms and counted the longest among nested jargon terms as a single term.
We treated words or special characters not constituting medical jargon as individual terms.
The sentences and terms are shown in Appendix \ref{appendix:A}.

\section{Methods}
We identified medical terms for each socio-demographic group based on the majority vote of Turkers in each group.
Using this as the ground-truth label, we assess medical term extraction performance from various sources: SciSpacy \citep{neumann-etal-2019-SciSpacy}, MedJEx \citep{kwon2022MedJEx}, and two OpenAI ChatGPT models.
In particular, we apply role-playing and in-context learning to observe their effects on performance.

\subsubsection*{Turkers}
We collected 270 responses from Turkers, including information about their primary language, frequency of health literacy reading, age, education (highest degree received), ethnicity, and gender.
Table 1 shows the distribution across these groups.

\begin{table*}[h]
    \centering
    \begin{tabular}{|c|c|c|}
    \hline
    Group Types & Group & Count \\
    \hline \hline
    \multirow{7}{*}{Age}  & 18-24 & 16 \\
        & 25-34 & 74 \\
        & 35-44 & 65 \\
        & 45-54 & 27 \\
        & 55-64 & 59 \\
        & 65-74 & 27 \\
        & 75+ & 2 \\
    \hline
    \multirow{5}{*}{Health literacy} & never & 133 \\
        & rarely & 95 \\
        & sometimes & 30 \\
        & often & 10 \\
        & always & 2 \\
    \hline
    \multirow{4}{*}{Education } & Low (Middle school or less) & 3 \\
        & HS (High school) & 87 \\
        & BA (Bachelor's degree) & 90 \\
        & MA (Postgraduate degree) & 90 \\
    \hline
    \multirow{3}{*}{Gender} & F (Female) & 149 \\
        & M (Male) & 120 \\ 
        & Other & 1 \\
    \hline

    \end{tabular}
    \caption{Socio-demographic distribution of 270 Turkers.}
    \label{table:data-distribution}
\end{table*}

To deal with groups with low number of people, we merged some groups.

\begin{enumerate}

    \item The age groups ``65-74'' and ``75+'' are merged into a single group labeled ``65+'' comprising 29 individuals.
    \item The age categories ``sometimes'', ``often'', and ``always'' are consolidated into a group denoted as ``$\ge$sometimes'' with a total of 42 participants.
    \item We combine the ``Low'' and ``HS'' groups and the ``$\le$HS'' group instead, which encompasses 90 individuals.
    \item Simplifying gender distinctions, we utilize only two groups: ``F'' for females and ``M'' for males.
\end{enumerate}

\subsubsection*{Medical Jargon Extractor Models}
We employed SciSpacy and MedJEx for extracting medical jargon terms.
Specifically, for SciSpacy, we used the ``en\_core\_sci\_scibert'' model.

\subsubsection*{ChatGPT with Role-Playing}
We extracted medical jargon using ChatGPT models with and without system messages that instruct the model to role-play a specific group.
In particular, we used two models, gpt-3.5-turbo (gpt-3.5-turbo-1106) and gpt-4 (gpt-4-0613) from OpenAI.

First, we gathered data with and without role-playing.
When role-playing was applied, we initiated a chat with one system message to instruct the models to role-play a specific socio-demographic group.
An example of such a system message is ``Your highest level of education is up to a high school diploma or equivalent. This encompasses scenarios from having no formal schooling or partial schooling to completing the standard education required before college or university, without pursuing further formal education.'' which is used for the ``$\le$HS'' group.
The system messages used for other groups can be found in the Appendix \ref{appendix:B}.

Subsequently, we gave one user message, which is
``Given the following sentence, please identify and enumerate all the medical terms present in a Python list format without any other descriptions: '' followed by a sentence from medical notes.
An example of the chat is shown in Figure 1.

For hyperparameters, we used ``max\_tokens''=256, ``top\_p''=1, ``frequent\_penalty''=0 and ``presence\_penalty''=0 with varying ``n'' and ``temperature'' between 0 and 1.
Here, ``frequent\_penalty'' and ``presence\_penalty'' influence the probability of output tokens based on the presence or frequency of the token in the generated output.
When these values are set to 0, the outputs are generated without consideration of the presence or frequency of a token.
``top\_p'' indicates the probability mass in nucleus sampling \citep{holtzman2019curious} where the model considers only the tokens comprising the top ``top\_p'' probability.
We used the default setting of ``top\_p'' equals to 1.0.
``max\_tokens'' is the maximum number of tokens to be generated.
Using higher ``temperature'' makes the output to be more random and ``n'' indicates the number of output to be generated.
When ``temperature'' is 0, we used one sample for each case by setting ``n'' = 1.
Additionally, we conducted the experiment with ``temperature'' of 0.2, 0.5, 0.7, and 1.0, generating 20 outputs for each setting.
We determined that ChatGPT selected a term to be a medical term based on the majority of the outputs.

\subsubsection*{ChatGPT with In-Context Learning}

We conduct experiments with in-context learning by partitioning the 20 sentences into five folds of four sentences each.
These four sentences, along with the ground truth based on the MTurkers' results, are then appended to the user message.

An example of user message using ground truth for one sentence is provided below:

``Given the following sentence, please identify and enumerate all the medical terms present in a Python list format without any other descriptions. \\

Input: This showed good patency of the vein graft with the absence of any residual valve and with the absence of any technical defect of the distal anastomosis and with prompt filling of the anterior tibial artery all the way down to the level of the foot and with retrograde filling of the posterior tibial artery through the pedal arch. \\
Output: [`distal anastomosis', `anterior tibial artery', `posterior tibial artery'] \\

Input: Pulmonary's impression was the patient has multiple pulmonary issues including a pulmonary embolism, has been therapeutic on her Coumadin, possible rheumatoid arthritis, induced lung injury versus interstitial lung disease, check CT scan with PE protocol, agrees with the IV steroids and nebulizers. \\
Output: ''\\

We conduct experiments with in-context learning with temperatures of 0.0 and 1.0, collecting one and four output for each fold respectively.

\subsection*{Data Cleaning}

For MedJEx and ChatGPT, we did not acquire the position of words to be predicted as a part of medical terms.
Consequently, for every obtained results, we marked all occurrences of each extracted medical term in each sentence and then re-collected the results by choosing terms where all of the words that consist a term are marked.

\section{Experiments}

\subsection*{Metrics}
Based on the Turkers' medical term extraction results, we labeled that a term is considered as a medical term following the majority vote in each group.
Then, we calculated F1 scores of each output's model for each group as shown in the following equations.

\begin{equation} \label{PRECISION}
Precision = \frac{\#\,of\,true\,positive\,medical\,terms}{\#\,of\,model\,predicted\,medical\,terms}
\end{equation}

\begin{equation} \label{RECALL}
Recall = \frac{\#\,of\,true\,positive\,medical\,terms}{\#\,of\,Turkers\,predicted\,medical\,terms}
\end{equation}

\begin{equation} \label{F1}
F_{1}\:score = 2\times\frac{Precision\times Recall}{Precision + Recall}
\end{equation}

Additionally, we report scores for the Wilcoxon signed-rank test comparing the calculated F1 scores with varying temperatures to show that the results with and without role-playing are statistically extracted from different distributions.

\subsection*{Results}

Table 2 shows the F1 scores of SciSpacy, MedJEx, and GPT models with and without in-context learning and with varying temperatures.

\begin{sidewaystable}
    \centering
    \begin{adjustbox}{width=1\textwidth}
        \begin{tabular}{|c|c|c|c|c|c|c|c|c|c|c|c|c|c|c|c|c|c|}
        \hline
        Model & Temp. & Role. & ICL & \multicolumn{3}{|c|}{Education} & \multicolumn{3}{|c|}{Health literacy} & \multicolumn{2}{|c|}{Gender} & \multicolumn{6}{|c|}{Age}   \\
        \hline
         &  &  &  & $\le$HS & BA & MA & never & rarely & $\ge$sometimes & F & M & 18-24 & 25-34 & 35-44 & 45-54 & 55-64 & 65+ \\
        \hline
        scispacy &  & No & No & 43.08 & 44.44 & 43.94 & 44.05 & 43.26 & 44.05 & 44.16 & 44.05 & 42.05 & 44.84 & 43.65 & 43.94 & 43.65 & 43.83 \\
        \hline
        medjex &  & No & No & 49.64 & 52.82 & 53.52 & 53.00 & 51.25 & 51.59 & 51.77 & 53.00 & 51.08 & 51.93 & 52.48 & 50.70 & 51.77 & 52.63 \\
        \hline
        gpt-3.5-turbo & 0.0 & No & No & 44.27 & 46.72 & 46.72 & 46.32 & 45.50 & 46.32 & 45.91 & 46.32 & 44.27 & 47.12 & 45.91 & 46.72 & 45.38 & 47.12 \\
        \hline
        gpt-3.5-turbo & 0.0 & Yes & No & 46.93 & 46.56 & 47.72 & 52.63 & 48.07 & 47.06 & 47.25 & 46.49 & 45.43 & 49.05 & 46.58 & 48.62 & 46.99 & 48.24 \\
        \hline
        gpt-3.5-turbo & 0.2 & No & No & 44.15 & 46.60 & 46.60 & 46.19 & 45.38 & 46.19 & 45.79 & 46.19 & 44.15 & 47.00 & 45.79 & 46.60 & 45.26 & 47.00 \\
        \hline
        gpt-3.5-turbo & 0.2 & Yes & No & 46.70 & 47.85 & 48.60 & 52.67 & 49.29 & 47.06 & 47.93 & 46.61 & 45.86 & 48.63 & 47.49 & 48.48 & 47.41 & 48.50 \\
        \hline
        gpt-3.5-turbo & 0.5 & No & No & 44.68 & 47.12 & 47.12 & 46.72 & 45.91 & 46.72 & 46.32 & 46.72 & 44.68 & 47.52 & 46.32 & 47.12 & 45.79 & 47.52 \\
        \hline
        gpt-3.5-turbo & 0.5 & Yes & No & 46.93 & 48.11 & 47.54 & 52.67 & 49.44 & 47.34 & 48.18 & 47.12 & 46.50 & 50.98 & 48.04 & 49.44 & 47.41 & 50.00 \\
        \hline
        gpt-3.5-turbo & 0.7 & No & No & 44.99 & 47.47 & 47.47 & 47.06 & 46.24 & 47.06 & 46.65 & 47.06 & 44.99 & 47.87 & 46.65 & 47.47 & 46.11 & 47.87 \\
        \hline
        gpt-3.5-turbo & 0.7 & Yes & No & 47.43 & 48.88 & 48.89 & 54.00 & 49.43 & 48.90 & 48.47 & 48.20 & 46.46 & 51.70 & 47.91 & 48.89 & 49.01 & 49.58 \\
        \hline
        gpt-3.5-turbo & 1.0 & No & No & 46.11 & 48.63 & 48.63 & 48.22 & 47.38 & 48.22 & 47.80 & 48.22 & 46.11 & 49.05 & 47.80 & 48.63 & 47.25 & 49.05 \\
        \hline
        gpt-3.5-turbo & 1.0 & Yes & No & 48.55 & 50.43 & 51.87 & 54.92 & 50.73 & 49.72 & 49.15 & 48.75 & 47.54 & 50.85 & 50.29 & 49.57 & 49.29 & 50.57 \\
        \hline
        gpt-3.5-turbo & 0.0 & No & Yes & 53.39 & 55.25 & 54.47 & 53.91 & \textbf{54.33} & 57.03 & 54.12 & 55.47 & 52.59 & 57.36 & 54.90 & 54.47 & 53.33 & 53.49 \\
        \hline
        gpt-3.5-turbo & 0.0 & Yes & Yes & 53.17 & 55.04 & 54.26 & 56.72 & \textbf{54.33} & 56.81 & 54.33 & 55.47 & 52.80 & 57.36 & 54.26 & 54.26 & 52.51 & 53.08 \\
        \hline
        gpt-3.5-turbo & 1.0 & No & Yes & 53.70 & 55.51 & 54.75 & 54.20 & 53.85 & 56.49 & 54.41 & 54.96 & 51.36 & 56.82 & 55.17 & 54.75 & 53.64 & 53.79 \\
        \hline
        gpt-3.5-turbo & 1.0 & Yes & Yes & 53.17 & 54.55 & 52.99 & \textbf{57.14} & 54.05 & 56.27 & 52.47 & 53.79 & 52.59 & 57.25 & 52.83 & 52.83 & 51.52 & 52.43 \\
        \hline
        gpt-4 & 0.0 & No & No & 46.28 & 48.17 & 48.17 & 47.77 & 47.49 & 48.29 & 47.89 & 47.77 & 45.74 & 49.09 & 47.37 & 48.17 & 47.89 & 48.56 \\
        \hline
        gpt-4 & 0.0 & Yes & No & 47.15 & 50.13 & 48.81 & 47.87 & 47.87 & 48.31 & 49.32 & 49.19 & 46.77 & 49.87 & 48.00 & 48.81 & 48.79 & 48.81 \\
        \hline
        gpt-4 & 0.2 & No & No & 46.19 & 48.58 & 48.58 & 48.19 & 47.40 & 48.19 & 47.79 & 48.19 & 46.19 & 48.97 & 47.79 & 48.58 & 47.79 & 48.97 \\
        \hline
        gpt-4 & 0.2 & Yes & No & 47.41 & 48.94 & 49.33 & 47.92 & 47.85 & 47.79 & 48.78 & 49.46 & 46.77 & 49.61 & 48.26 & 48.42 & 48.26 & 49.47 \\
        \hline
        gpt-4 & 0.5 & No & No & 46.19 & 48.58 & 48.58 & 48.19 & 47.40 & 48.19 & 47.79 & 48.19 & 46.19 & 48.97 & 47.79 & 48.58 & 47.79 & 48.97 \\
        \hline
        gpt-4 & 0.5 & Yes & No & 47.80 & 49.46 & 48.94 & 48.69 & 47.72 & 48.19 & 49.45 & 49.34 & 47.06 & 49.74 & 47.75 & 49.07 & 48.66 & 49.86 \\
        \hline
        gpt-4 & 0.7 & No & No & 46.07 & 48.45 & 48.45 & 48.06 & 47.27 & 48.06 & 47.67 & 48.06 & 46.07 & 48.84 & 47.67 & 48.45 & 47.67 & 48.84 \\
        \hline
        gpt-4 & 0.7 & Yes & No & 47.93 & 49.34 & 49.33 & 48.69 & 47.34 & 48.06 & 49.32 & 49.46 & 46.52 & 49.74 & 48.81 & 48.81 & 49.06 & 49.87 \\
        \hline
        gpt-4 & 1.0 & No & No & 45.83 & 48.21 & 48.21 & 47.81 & 47.03 & 47.81 & 47.42 & 47.81 & 45.83 & 48.59 & 47.42 & 48.21 & 47.42 & 48.59 \\
        \hline
        gpt-4 & 1.0 & Yes & No & 48.07 & 50.00 & 49.46 & 49.08 & 47.49 & 48.44 & 49.18 & 49.60 & 46.81 & 49.74 & 48.94 & 49.47 & 49.32 & 49.87 \\
        \hline
        gpt-4 & 0.0 & No & Yes & 53.79 & 55.56 & 54.81 & 54.28 & 53.93 & 56.51 & 54.48 & 55.02 & 51.52 & 56.83 & \textbf{55.22} & 54.81 & 53.73 & 53.87 \\
        \hline
        gpt-4 & 0.0 & Yes & Yes & \textbf{54.55} & \textbf{56.82} & \textbf{55.76} & 54.76 & 53.03 & \textbf{57.25} & \textbf{55.43} & \textbf{58.37} & 52.38 & \textbf{58.65} & 54.83 & \textbf{54.96} & \textbf{54.28} & 53.38 \\
        \hline
        gpt-4 & 1.0 & No & Yes & 53.38 & 55.15 & 54.41 & 53.87 & 53.53 & 56.09 & 54.07 & 54.61 & 51.13 & 56.41 & 54.81 & 54.41 & 53.33 & 53.48 \\
        \hline
        gpt-4 & 1.0 & Yes & Yes & 53.93 & 55.68 & 55.52 & 55.91 & 53.18 & 56.62 & 54.61 & 56.12 & \textbf{53.76} & 54.81 & 54.95 & 53.95 & 51.85 & \textbf{55.89} \\
        \hline
    \end{tabular}
    \end{adjustbox} 
    \caption{F1 score between each model and respondent results. Temp and Role indicate temperature and role-playing respectively. For ChatGPT instances, 20 outputs are used except for when the temperature is 0.0. The best performing scores for each group are highlighted in bold.}
    \label{table:data-distribution}
\end{sidewaystable}

\subsubsection*{Effect of role-playing}

We investigated whether role-playing in LLMs is effective for reflecting the socio-demographic background of individuals.
As illustrated in Figure 2, applying role-playing increased the macro F1 scores for both models, with and without in-context learning.
Notably, role-playing was more effective when in-context learning was not applied.

We further repeated the experiments with varying temperatures of 0.0, 0.2, 0.5, 0.7, and 1.0 without applying in-context learning and observed that the results improved with role-playing in 133 out of 140 cases (95\%).
The results of Wilcoxon signed-rank test on the results are shown in Table 3.
Notably, an increase in the macro F1 score is evident across all instances except for the cases when in-context learning was applied to gpt-3.5-turbo.
Additionally, the p-values from the Wilcoxon signed-rank test being less than 0.05 show that the F1 scores in each group likely follow distinct distributions in all cases where in-context learning was not applied. 
The higher p-values and less increase in macro F1 scores associated with in-context learning show that providing specific examples impact of the roles assigned to the models.

\begin{table*}
    \centering
    \begin{tabular}{|c|c|c|c|c|c|c|c|}
        \hline
        Model & Temp. & N & ICL & Macro F1 w/o Role & Macro-F1 w/ Role & Diff. & p-value \\
        \hline
        gpt-3.5-turbo & 0.0 & 1 & No & 46.06 & 47.64 & 1.58$^{\ddag}$ & 0.000122 \\
        \hline
        gpt-3.5-turbo & 0.0 & 4 & Yes & 54.54 & 54.51 & -0.02 & 0.099351 \\
        \hline
        gpt-3.5-turbo & 0.2 & 20 & No & 45.94 & 48.02 & 2.08$^{\ddag}$ & 6.1e-05 \\
        \hline
        gpt-3.5-turbo & 0.5 & 20 & No & 46.46 & 48.5 & 2.03$^{\ddag}$ & 6.1e-05 \\
        \hline
        gpt-3.5-turbo & 0.7 & 20 & No & 46.8 & 49.02 & 2.22$^{\ddag}$ & 6.1e-05 \\
        \hline
        gpt-3.5-turbo & 1.0 & 20 & No & 47.96 & 50.02 & 2.06$^{\ddag}$ & 0.000122 \\
        \hline
        gpt-3.5-turbo & 1.0 & 20 & Yes & 54.51 & 53.79 & -0.71 & 0.063721 \\
        \hline
        gpt-4 & 0.0 & 1 & No & 47.76 & 48.39 & 0.63$^{\dag}$ & 0.006714 \\
        \hline
        gpt-4 & 0.0 & 4 & Yes & 54.58 & 55.21 & 0.63$^{*}$ & 0.030151 \\
        \hline
        gpt-4 & 0.2 & 20 & No & 47.97 & 48.31 & 0.34$^{*}$ & 0.030151 \\
        \hline
        gpt-4 & 0.5 & 20 & No & 47.97 & 48.57 & 0.6$^{*}$ & 0.01315 \\
        \hline
        gpt-4 & 0.7 & 20 & No & 47.85 & 48.63 & 0.78$^{\dag}$ & 0.002865 \\
        \hline
        gpt-4 & 1.0 & 20 & No & 47.6 & 48.86 & 1.26$^{\ddag}$ & 0.000122 \\
        \hline
        gpt-4 & 1.0 & 20 & Yes & 54.17 & 54.7 & 0.53 & 0.094604 \\
        \hline
    \end{tabular}
    \caption{Macro F1 scores and p-value of Wilcoxon Sign Test with various settings using ChatGPT. 'Temp.' indicates the temperature, 'N' indicates number of samples per sentence, 'ICL' indicates whether in-context learning was applied, '$\Delta$' shows the difference in macro f1 score after applying role-playing, and p-value indicates the p-value of Wilcoxon Sign Test. $^*$, $^\dag$, and $^\ddag$ indicate the p-value of Wilcoxon Sign Test being smaller than 0.05, 0.01, and 0.001, respectively.}
    \label{table:data-distribution}
\end{table*}

\subsubsection*{Comparison with previous models}

To compare with previous models, we assessed the F1 scores of ChatGPT in relation to those of SciSpacy and MedJEx.
Overall, without in-context learning, ChatGPT demonstrated superior performance compared to SciSpacy but did not reach the performance level of MedJEx.
For example, with the default temperature of 1.0, the macro F1 scores were 42.86 for SciSpacy and 50.92 for MedJEx, while ChatGPT achieved scores between 46.45 and 47.83.
However, when applying role-playing with in-context learning, GPT-4 surpassed MedJEx, the state-of-the-art model, achieving a macro average F1 score of 51.42.
This trend was consistent across various temperature settings.


\begin{figure*}[htbp]
    \centering
    \includegraphics[width=0.5\linewidth]{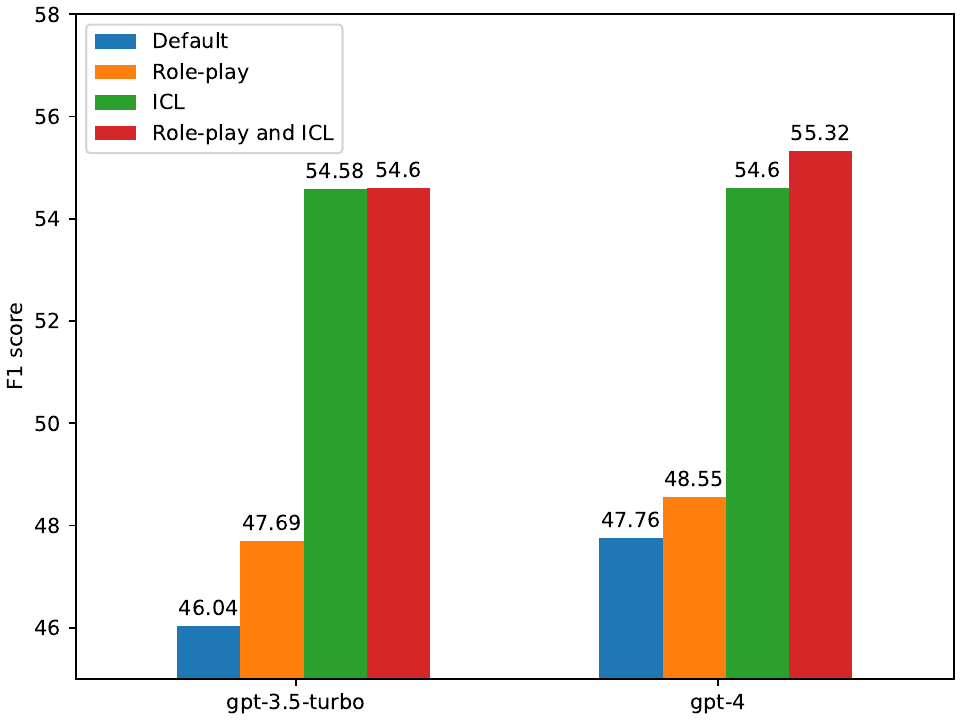}
    \caption{Macro F1 scores using gpt-3.5-turbo and gpt-4 with the temperature of 0.0.}
    \label{fig:MCC_difference}
\end{figure*}

\subsubsection*{Comparison by group}

Figure 3 shows the macro F1 scores for each group without in-context learning when temperature was 1.0.
Applying role-playing increased the performance in all cases with most significant changes in F1 scores for GPT-3.5 and GPT-4 being in Health literacy and Gender respectively.



\begin{figure*}[htbp]
    \centering
    \includegraphics[width=0.5\textwidth]{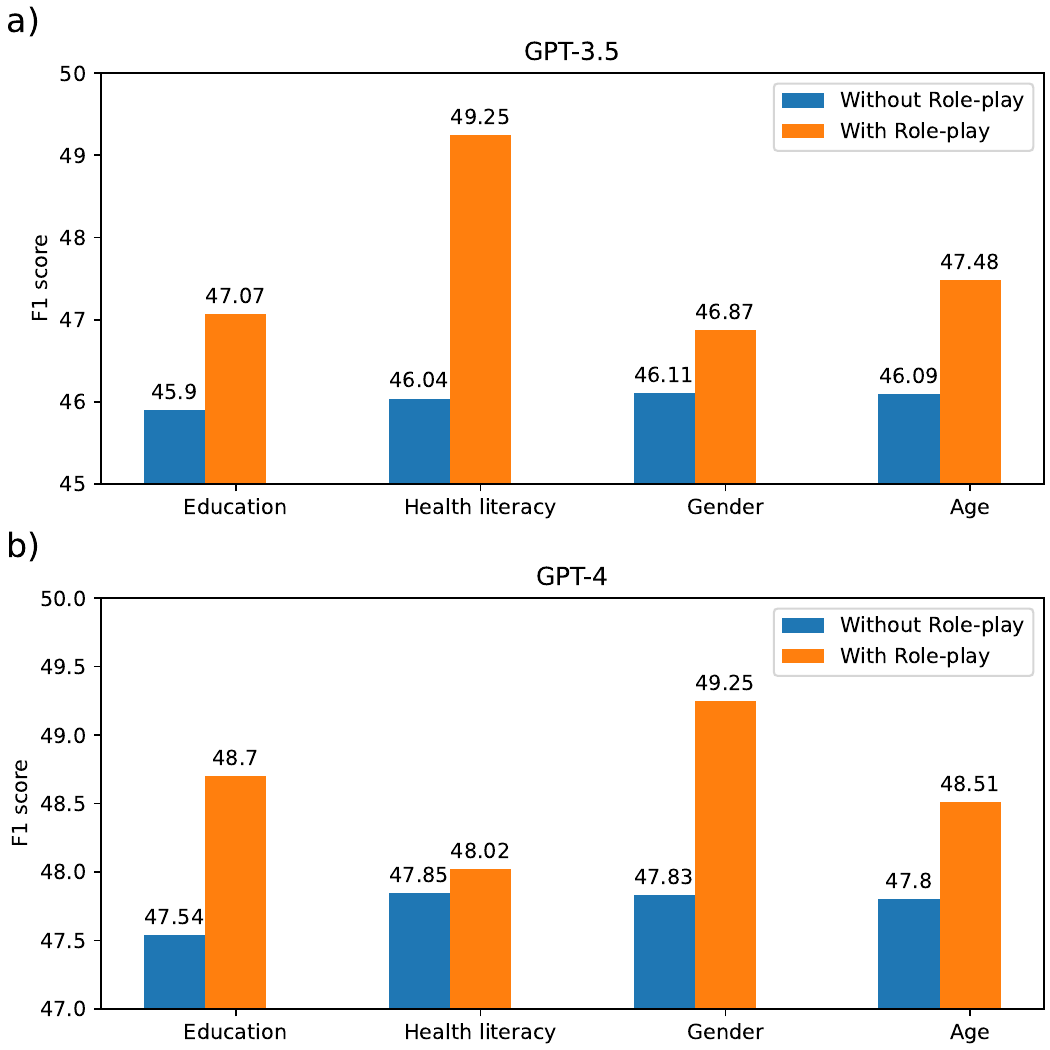}
    \caption{Macro F1 scores with and without role-play for each socio-demographic factor with the temperature of 0.0.}
    \label{fig:MCC_difference}
\end{figure*}


\section{Discussion}

\subsection*{The task of medical term extraction}

``Medical term'' (Medical terminology) refers to the words and language used specifically in the medical and health fields.
Identifying medical terms may depend on both common sense and background knowledge.
Consider the medical term ``lumbar spinal surgery''.
Even without knowing the meaning of ``lumbar spinal'', it appears to be adverbs related to surgery, hence likely medical terms.
Similarly, ``cine catheterization'' in the phrase ``cine catheterization was done by Dr.[NAME]'' can be chosen as a medical term without medical expertise since the context suggests that it is an action performed by a doctor in response to a patient's health status.
Likewise, identifying medical terms could be intuitive based on medical root words.
For example,``cardio-'' and ``vascular'' has meanings of ``pertaining to heart'' and ``pertaining to conveyance or circulation of fluids'', respectively in etymology.
Knowing these medical root words, one could easily identify that cardiovascular exercise is any activity related to heart rate pumping.

On the other hand, the task may be challenging since a word can be considered as a medical term depending on the context.
For example, the word ``run'' can simply mean to move at a speed faster than a walk.
However, it can also be a medical jargon when a doctor says ``We need to run some test'', by meaning to ``perform or conduct medical tests''.

In short, while largely being an intuitive task, the task of medical term extraction could be difficult without proper background knowledge and due to the ambiguity of the definition of ``medical term''.

\subsection*{Merits}

Personalization in medical term extraction can be used to improve communication between various people.
Identifying medical terms that each person find difficult can help improve the comprehension of readers by avoid using medical jargons or adding additional explanations for difficult medical terms.
Similarly, it can be adapted to downstream medical AI application systems such as NoteAid \citep{polepalli2013improving}, medical concept linking system, or chatbot-based self-diagnosis systems \citep{you2020self}.
The improved communication will, in turn, help patients to be self-aware of their own disease \citep{adams2010improving} and prevent physicians' burnouts \citep{aaronson2019training}.

Furthermore, our results show the potential of ChatGPT being used in the bio-medical domain to personalize documents such as EHR.
Medical doctors with different specialties, diverse level of medical students, nurses and patients with varying socio-demographic backgrounds, and pharmacists could benefit from such results by improved comprehension in medical documents which could otherwise be only understood by peer expert doctors.

\subsection*{Limitations}

Our experiments were conducted with only 20 sentences.
We found that out of 20 sentences, the majority vote for extracted medical terms for the 14 different socio-demographic background that we investigated differed only in 31 terms.
This lead to insignificant changes in F1 scores in our results.
For instance, sentences 3, 4, 8, 12, 15, and 20 (Appendix \ref{appendix:A}) gave no meaningful results for the effect of role-playing because people from all socio-demographic background that we investigated gave the same results.



\subsection*{Future Works}

For the future work, we plan to leverage the capabilities of role-playing in ChatGPT for practical uses, including customizing Electronic Health Record (EHR) notes, developing systems for linking medical concepts, and creating chatbot-based self-diagnosis tools.



\section{Conclusion}

Our analysis of ChatGPT's performance in medical term extraction shows that its performance can outperform than the state-of-the-art model without domain-specific training or fine-tuning in the downstream task.
Moreover, we have shown that ChatGPT generates better prediction when role-playing is applied to predict Turkers' result based on their sociodemographic background.
These findings show the potential for leveraging role-playing in large language models to personalize information within the biomedical domain.







\makeatletter
\renewcommand{\@biblabel}[1]{\hfill #1.}
\makeatother

\bibliography{main}

\newpage
\appendix
\section{Appendix}

\subsection*{Appendix A: Details On the Experimental Data}
\label{appendix:A}

This section outlines the 20 sentences utilized in the experiment. Herein, individual words are represented by spaces. However, multi-word expressions, like `anterior tibial artery,' are demarcated with `[]' symbols. To conform HIPAA regulation, personal information has been de-identified as AGE, NAME, INSTITUTION.




\textbf{Sentence 1:}
This showed good patency of the [vein graft] with absence of any [residual valve] and with absence of any [technical defect] of the [distal anastomosis] and with [prompt filling] of the [anterior tibial artery] all the way down to the level of the foot and with [retrograde filling] of the [posterior tibial artery] through the [pedal arch] .

\textbf{Sentence 2:}
FINDINGS : There are at least 3 distinct areas of [restricted diffusion] involving the left [occipital pole] , the left [medial temporal lobe] , and the left [basal ganglia region] possibly overlying the [posterior limb] of the left [internal capsule] .

\textbf{Sentence 3:}
In particular , he has had no symptoms of adenopathy or splenomegaly , [exertional chest pain] or orthopnea , dysphagia or [abdominal pain] , change in [bowel habits] or blood in the stool , dysuria , hematuria , abnormal bruising or bleeding , fevers or [drenching sweats] , or unexplained weight loss .

\textbf{Sentence 4:}
The patient has been ordered a dose of IV Ferrlecit for her [chronic iron deficiency] due to [GI blood loss] from AVM 's and will receive that prior to discharge , and as long as she remains [hemodynamically stable] , will be discharged home to INSTITUTION today .

\textbf{Sentence 5:}
Differential is somewhat broad and includes [acute intermittent porphyria] as well as nephrolithiasis , [gastric outlet obstruction] , esophagitis , pancreatitis , [intestinal motility disorder] , [factitious disorder] , or even [irritable bowel syndrome] .

\textbf{Sentence 6:}
Evaluation of the [lung parenchyma] is [markedly compromised] by [respiratory motion] with [mild upper lobe predominant ground glass opacity] and a 1 cm [inflammatory nodule] in the [posterior basal] left [lower lobe] . 

\textbf{Sentence 7:}
TECHNIQUE : [Helical CT images] of the thoracic and [lumbar spine] were obtained in [2.5 mm collimation] without the use of [IV contrast] and reconstructed using [bone algorithm] , subsequently coronal and [sagittal reformations] were performed as well .

\textbf{Sentence 8:}
[Metastatic colon cancer] to liver ; [status post] right [hepatic lobectomy] , [open radiofrequency ablation] of [left lobe lesion] , aspiration of [left lobe hepatic cyst] , [liver ultrasound] with doppler , [intraoperative cholangiogram] , [open cholecystectomy] , dated Dec 19 2007 ; [uncomplicated postoperative recovery] .

\textbf{Sentence 9:}
[DISCHARGE MEDICATIONS] : Reviewed ( Tylenol 650 MG PO [Q 6 HR] ampicillin 2 GM IV to complete 14 days Oscal 2 tab PO BID soma 350 MG PO BID Cardura 2 MG PO BID [ferrous sulfate] 325 MG PO daily [milk of magnesia] , 30 mL PO daily PRN ( constipation ) senna 1 tab PO QHS ) .

\textbf{Sentence 10:}
DM - pt [blood glucose] much more controlled when on Lantus as scheduled , coupled with Humalog [sliding scale] as pt is on at home - pt admitted to not being totally compliant with [insulin dosing] at home , particularly more so with the Humalog than with the Lantus - our concern at this point is whether the pt is actually drawing up her insulin properly at home when she does administer it - may consider adding an [oral agent] such as metformin to decrease insulin needs at home .

\textbf{Sentence 11:}
The patient was scheduled for [lumbar spinal surgery] , but because of a concern about heart condition , [cine catheterization] was done by Dr. NAME today showing severe [coronary artery disease] including [proximal long] 60 \% [RCA lesion] , 90 \% lesions affecting the LAD and [diagonal branch] , and 70 \% lesion at the OM1 of the [circumflex artery] with LVEF 60 \% to 70 \% .

\textbf{Sentence 12:}
The following result was found : Antigen / antibody [usual reactivity result] [CD34 progenitor cells] , [rare positive cells] . [Endothelial cells] , other [medical necessity justification] for the [immunohistochemical stains] that were needed in addition to the [flow cytometric immunophenotypic studies] for the best diagnosis possible is as follows : The [flow cytometric studies] did not define the precise nature of all the [cellular elements] of concern in this specimen .

\textbf{Sentence 13:}
[PAST MEDICAL HISTORY] : Notable for hypertension ; [fecal incontinence] ; [chronic diarrhea] ; [feeding tube] ; achalasia with reflux ; [esophageal stricture] ; [hiatal hernia] , [status post repair] ; [iron - deficiency anemia] ; [chronic headache] ; [B12 deficiency] ; [sleep disorder] ; and sarcoidosis .

\textbf{Sentence 14:}
[ASSESSMENT AND PLAN] : This is a [AGE-year-old] man with a history of [bipolar disorder] , [alcohol abuse] , [coronary artery disease] , [status post] [coronary artery bypass grafting] , and [chronic obstructive pulmonary disease] who presents with 3 months of decreased appetite , [lower extremity] weakness , and weight loss .

\textbf{Sentence 15:}
At baseline , the patient was seen to be in [sinus rhythm] with [sinus cycle length] of 900 milliseconds , [P-wave duration] of 95 milliseconds , PR interval of 140 milliseconds , [QRS duration] of 87 milliseconds , [QT interval] of 404 milliseconds , [AH interval] of 60 milliseconds , and [HV interval] of 38 milliseconds .

\textbf{Sentence 16:}
Echocardiogram showed LVEF of 60 \% ; normal left [ventricular systolic function] with evidence of [diastolic dysfunction] ; [mitral annular calcification] with mild [mitral regurgitation] ; [aortic sclerosis] ; [aortic root dilations] , multiple ; and [tricuspid regurgitation] .

\textbf{Sentence 17:}
Pulmonary 's impression was the patient has multiple pulmonary issues including a [pulmonary embolism] , has been therapeutic on her Coumadin , possible [rheumatoid arthritis] , [induced lung injury] versus [interstitial lung disease] , check CT scan with [PE protocol] , agrees with the IV steroids and nebulizers .

\textbf{Sentence 18:}
IMAGING : He had a [CT scan] on Aug 28 2008 of the head and neck , which reveals [operative changes] but significant tumor occurrence with [extensive involvement] of the left [masticator space] , parotid , [external ear] , temporal scalp . [Temporalis masseter] and [lateral pterygoid muscles] with possible [orbital involvement] .

\textbf{Sentence 19:}
This [AGE-year-old] female with a [past medical history] of acromegaly , [status post] [growth hormone-secreting tumor] which was resected in 2005 with a history of [pan hypopituitarism] , hypercholesterolemia , DI , [renal insufficiency] , [sixth nerve palsy] , and a [VP shunt placement] in 2005 , presents with headache .

\textbf{Sentence 20:}
CT of the abdomen and pelvis from Oct 26 shows progression of left [rectus sheath hematoma] caudally to the level of the [pubic symphysis] , [trace perihepatic fluid] , stable left [common iliac] aneurysm , and [stable intraperitoneal hemorrhage] in the [deep pelvis] .

\subsection*{Appendix B: ChatGPT System Messages}
\label{Appendix:B}

\subsubsection*{Education}
``$\le$HS'': ``Your highest level of education is up to a high school diploma or equivalent. This encompasses scenarios from having no formal schooling or partial schooling, to completing the standard education required before college or university, without pursuing further formal education.'' \\
``BA'': ``Your highest level of education is a Bachelor's degree from a college or university. You've undergone tertiary education and have specialized in a particular field or subject area.'' \\
``MA'': ``Your highest level of education is a Master's degree. You've not only completed your Bachelor's degree but have also undertaken advanced studies in a specific field, gaining deeper knowledge and perhaps conducting some level of research.'' 

\subsubsection*{Health literacy}
``never'': ``You less than rarely engaged with health literacy materials and are unfamiliar with basic health information and medical terms.'' \\
``rarely'': ``You rarely come across health literacy materials and only occasionally glance at health-related information.'' \\
``$\ge$sometimes'': ``Your interaction with health literacy materials is frequent, indicating that you engage with them sometimes, often, or always in your everyday life.'' 

\subsubsection*{Age}

``18'': ``You are a person in the age range of 18-24.'' \\
``25'': ``You are a person in the age range of 25-34.'' \\
``35'': ``You are a person in the age range of 35-44.'' \\ 
``45'': ``You are a person in the age range of 45-54.'' \\
``55'': ``You are a person in the age range of 55-64.'' \\
``65/75'': ``You are a person above the age of 65.'' 

\subsubsection*{Gender}
``M'': ``You are a man. You identify your gender as a male.'' \\
``F'': ``You are a woman. You identify your gender as a female.'' 



\end{document}